\documentclass[final,12pt,3p]{elsarticle}

\usepackage{graphicx}
\usepackage{amssymb}
\usepackage{amsthm}
\usepackage{amsmath}
\usepackage{algorithm}
\usepackage{algorithmic}
\usepackage{tikz}
\usepackage{lineno}
\usepackage{hyperref}
\usepackage{verbatim}

\usepackage{caption}
\usetikzlibrary{shapes}
\usepackage{subcaption}
\usepackage{geometry}
\usepackage{mathrsfs} 
\usepackage{array}
\usepackage{multirow}
\usepackage{booktabs} 
\usepackage{adjustbox}
\usepackage[title]{appendix}
\usepackage{xcolor}

\biboptions{sort&compress}
\hypersetup{
colorlinks=true,
linkcolor=red,
filecolor=green,
urlcolor=green,
anchorcolor=green,
citecolor=cyan,
pdftitle={Overleaf Example},
pdfpagemode=FullScreen,
}
\newtheorem{lemma}{Lemma}
\newtheorem{definition}{Definition}
\newtheorem{remark}{Remark}

\newtheorem{proposition}{Proposition}

\newcommand{\C}{\mathcal{C}}
\newcommand{\Dat}{\mathcal{D}^{\alpha}_{t}}
\newcommand{\Ia}{\mathcal{I}^{\alpha}}
\renewcommand{\Re}{\operatorname{Re}}

 \usepackage[figcolor=white]{todonotes}

\begin{document}
\begin{frontmatter}

\title{Identifying Memory Effects in Epidemics via a Fractional SEIRD Model and Physics-Informed Neural Networks}



\author{Achraf Zinihi}
\ead{a.zinihi@edu.umi.ac.ma}

\address{University of Wuppertal, Applied and Computational Mathematics,\\
Gaußstrasse 20, 42119 Wuppertal, Germany}

\address{Moulay Ismail University of Meknes, FST Errachidia, Department of Mathematics,\\ AMNEA Group, Errachidia 52000, Morocco}


\begin{abstract}
%
%

We develop a physics-informed neural network (PINN) framework for parameter estimation in fractional-order SEIRD epidemic models. By embedding the Caputo fractional derivative into the network residuals via the L1 discretization scheme, our method simultaneously reconstructs epidemic trajectories and infers both epidemiological parameters and the fractional memory order $\alpha$.
The fractional formulation extends classical integer-order models by capturing long-range memory effects in disease progression, incubation, and recovery. \\
Our framework learns the fractional memory order $\alpha$
as a trainable parameter while simultaneously estimating the epidemiological rates $(\beta, \sigma, \gamma, \mu)$. 
A composite loss combining data misfit, physics residuals, and initial conditions, with constraints on positivity and population conservation, ensures both accuracy and biological consistency. Tests on synthetic Mpox data confirm reliable recovery of 
$\alpha$ and parameters under noise, while applications to COVID-19 show that optimal $\alpha \in (0, 1]$
 captures memory effects and improves predictive performance over the classical SEIRD model.\\
This work establishes PINNs as a robust tool for learning memory effects in epidemic dynamics, with implications for forecasting, control strategies, and the analysis of non-Markovian epidemic processes.
\end{abstract}


\journal{} 



\end{frontmatter}


\section{Introduction}\label{S1}

The global spread of infectious diseases such as the Black Death (1347–1351 in Europe, Asia, and North Africa), smallpox (16th–20th century, eradicated the Americas in 1980), cholera (seven waves from 1817 to 1975), H1N1 influenza (2009–2010, swine flu), Ebola (notably 2014–2016 in West Africa), and COVID-19 (2019--present, caused by SARS-CoV-2) has highlighted the importance of reliable epidemic models for understanding disease dynamics and guiding public health interventions \cite{Britannica2025, CDC2025, Hu2016, Zimmer2009, Zinihi2025Koopman, WHO2016, CDC2025COVID}. 
Traditional compartmental models, such as the classical SI, SIR, SEIR and SEIRD frameworks, have long served as fundamental tools for capturing the transmission, progression, and recovery processes in populations \cite{Balderrama2024, Zinihi2025S, Sun2023, Ghersheen2019, Duan2014, Zinihi2025A, Hethcote2000}. 
These models are widely used due to their simplicity and analytical tractability. However, they often fail to fully capture the real-world epidemic dynamics, such as the long-term memory effects and anomalous diffusion observed in disease spreading.

To address these limitations, researchers have increasingly turned to fractional-order epidemic models. These models incorporate memory and hereditary effects into disease dynamics by replacing classical integer-order derivatives with fractional operators  \cite{Adak2024, Adu2024, Zinihi2025FDE, Barros2021, Diethelm2010, Barman2025, Zinihi2025MM}. 
The fractional-order parameter, usually denoted by $\alpha$, characterizes the degree of memory in the system. When $\alpha=1$, the model reduces to the standard integer-order case. 
When $0 < \alpha < 1$, the model reflects subdiffusive, history-dependent dynamics. 
Several studies have demonstrated that fractional epidemic models can provide more accurate fits to epidemiological data and reveal dynamics not captured by classical models \cite{Zinihi2025MM, Almuzini2025, MaurciodeCarvalho2021}.
Readers seeking a more in-depth discussion are referred to the following comprehensive studies \cite{Li2018, Feng2024, Maione2023, Fang2020, Lenka2025, Verma2023, Caputo2017, Pandey1993, Antony2026, Khalid2023}. These studies provide extensive coverage of fractional calculus applications.

Despite these advantages, applying fractional epidemic models presents several challenges. Estimating the memory order $\alpha$ from real epidemic data is particularly challenging. 
It is often nonintuitive and highly sensitive to noise or data sparsity. Furthermore, it is strongly correlated with other epidemiological rates. 
Furthermore, classical fitting approaches based on least-squares optimization require repeated numerical simulations of the fractional model. 
These simulations are computationally demanding due to the nonlocal nature of the Caputo derivative \cite{Zinihi2025OC, Diethelm2012, Naik2021, Kumar2021}. 
Conversely, purely data-driven machine learning approaches often ignore the underlying epidemiological structure, leading to biologically inconsistent outcomes, such as negative compartment sizes or loss of population conservation. 
These challenges motivate developing hybrid approaches that integrate the mechanistic structure of epidemic models with the flexibility of modern data-driven methods.

In this regard, \textit{physics-informed neural networks} (PINNs) are a promising alternative. By embedding the governing equations into the neural network training process, PINNs enforce physical consistency while fitting observational data simultaneously
\cite{Raissi2019, Heldmann2023, Berkhahn2022}. 
This makes them particularly attractive for epidemiological models, where constraints such as positivity, conservation, and dynamical consistency are essential. 
While recent works have successfully applied PINNs to epidemic modeling in the integer-order setting \cite{Berkhahn2022, Han2024, Millevoi2024, Hwang2022}, the integration of fractional-order operators into PINNs remains largely unexplored.

In this study, we present a PINN-based framework for estimating the memory order $\alpha$ and, optionally, key epidemiological parameters, such as transmission, recovery, and mortality rates, in fractional SEIRD models. 
Our methodology uses the L1 discretization scheme for the Caputo derivative to represent fractional dynamics in the neural network loss function. We validate our approach on synthetic datasets where the ground-truth parameters are known and on real datasets from multiple regions in Europe. 
In addition to demonstrating the feasibility of estimating $\alpha$ from data, we examine the identifiability and interpretability of fractional epidemic dynamics compared to classical integer-order models. 
The main contributions of this study are:
\begin{itemize}
\item We develop a fractional SEIRD model with Caputo derivatives that incorporates memory effects into epidemic dynamics.

\item We design a PINN framework that can handle fractional operators via the L1 scheme and enforce epidemiological constraints.

\item We demonstrate that the method can recover the fractional memory order $\alpha$ and epidemiological parameters from synthetic and real epidemic datasets.

\item We perform uncertainty and identifiability analyses to highlight the role of memory effects in the spread of epidemics and their implications for forecasting.
\end{itemize}

This work lies at the intersection of fractional calculus, epidemiological modeling, and machine learning. It offers a novel perspective on data-driven parameter estimation in epidemic systems. The rest of the paper is organized as follows. 
Section~\ref{S2} provides the necessary preliminaries on fractional calculus and neural networks. 
Section~\ref{S3} introduces the fractional SEIRD model and presents its well-posedness analysis. 
The proposed methodology, including the PINN framework, the discretization of the Caputo derivative, the parameter estimation strategy, the loss function, and the  training algorithm, is detailed in Section~\ref{S4}. 
Section~\ref{S5} describes the experiments on both synthetic and real epidemic datasets and discusses the obtained results. Finally, Section~\ref{S6} concludes the paper and outlines directions for future work.

\section{Preliminaries}\label{S2}

This section establishes the mathematical foundation of our study. Its purpose is to enable readers, especially those unfamiliar with fractional calculus, to understand the subsequent model description. 
In this work, we adopt the Caputo fractional derivative of order $\alpha \in (0, 1)$, which is well-suited for initial value problems.
Let $T > 0$ be a finite time horizon, and let $f\colon[0,T] \to \mathbb{R}$ be a sufficiently smooth function.

First, we recall the Gamma function, which generalizes the factorial function to complex arguments and often appears  in the formulation of fractional operators.
\begin{definition}[See {\cite[p. 24]{kilbas2006}}]\label{D1}
Let $\omega \in \mathbb{C}$ such that $\Re(\omega) > 0$. 
The Gamma function is given by
\begin{equation*}
\Gamma(\omega):=\int_{0}^{+\infty} e^{-\tau} \tau^{\omega-1} d \tau.
\end{equation*}
\end{definition}

The Gamma function can be used to introduce the fractional integral operator, which generalizes the concept of repeated integration to include non-integer orders.
\begin{definition}[See {\cite[p. 69]{kilbas2006}}]\label{D2}
The fractional integral operator with base point $0$ is written as
\begin{equation*}
\Ia f(t) = \frac{1}{\Gamma(\alpha)} \int_{0}^{t} (t - \tau)^{\alpha-1} f(\tau) \,d \tau.
\end{equation*}
\end{definition}

For completeness, we recall the definitions of the left- and right-sided Caputo derivatives, since they play a central role in the formulation and analysis of fractional differential equations. 

\begin{definition}[See {\cite[p. 91]{kilbas2006}}]\label{D3}
The left and right fractional derivatives in the Caputo sense of $f$ with base point $0$ and $T$, respectively, are defined by
\begin{equation}\label{E2.1}
{ }^\C \Dat f(t) 
= \frac{1}{\Gamma(1-\alpha)} \int_{0}^{t} (t - \tau)^{-\alpha} f^{\prime}(\tau) \, d \tau,
\end{equation}
and 
\begin{equation*}
{ }^\C_{T} \Dat f(t) 
= \frac{-1}{\Gamma(1-\alpha)} 
\int_{t}^{T} (\tau - t)^{-\alpha} f^{\prime}(\tau) \, d \tau.
\end{equation*}
\end{definition}

\begin{remark}\label{R1}
As $\alpha$ approaches 1 from below, the case  reduces to the classical first-order derivative, recovering the memoryless ordinary differential equation (ODE) framework.
\end{remark}

We now recall a useful identity linking the Caputo derivative and the fractional integral, which will be used later in the analysis of our model.
\begin{lemma}[See {\cite[p. 96]{kilbas2006}}]\label{L1}
One has 
\begin{equation*}
\Ia \left( { }^\C 
\Dat f(t)\right) = f(t) - f(0).
\end{equation*}
\end{lemma}

\section{Model Description}\label{S3}

This section presents the mathematical framework used to describe the transmission dynamics of infectious diseases. 
We adopt an SEIRD compartmental model, which partitions the population into five epidemiological groups. 
This structure provides a flexible basis for incorporating fractional-order dynamics that capture memory effects in disease progression.

\subsection{Fractional SEIRD Model}\label{S3.1}

The SEIRD model is a widely adopted compartmental structure in mathematical epidemiology because it captures the fundamental stages of disease progression from exposure to removal. 
In this formulation, the total population is divided into five epidemiological classes: susceptible individuals $S(t)$, who are at risk of infection; exposed individuals $E(t)$, who have been infected but are not yet infectious; infectious individuals $I(t)$, who can transmit the disease to others; recovered individuals $R(t)$, who have acquired immunity; and deceased individuals $D(t)$, who have died due to the disease.

The transmission dynamics are governed by four key epidemiological parameters: 
\begin{center}
\begin{minipage}[t]{.35\textwidth}
\begin{center}
\begin{itemize}
\item[$(i)$] the infection rate $\beta$,
\item[$(iii)$] the recovery rate $\gamma$,
\end{itemize}
\end{center}
\end{minipage}
\hfill
\begin{minipage}[t]{.64\textwidth}
\begin{itemize}
\item[$(ii)$] the incubation rate $\sigma$,
\item[$(iv)$] the disease-induced mortality rate $\mu$.
\end{itemize}
\end{minipage}
\end{center}
To incorporate memory effects and long-range temporal dependencies, we extend the model using a Caputo fractional derivative of order $\alpha \in (0, 1]$. 
This generalization allows the current rate of change in each compartment to depend on both the present state and the history of the process. 
Thus, it captures the nonlocal character of epidemic spread.

Mathematically, the resulting fractional-order SEIRD system can be written as
\begin{equation}\label{E3.1}
\left\{\begin{aligned}
{ }^{\C}\Dat S(t) &= - \beta \, \frac{S(t) I(t)}{N_L(t)}, \\
{ }^{\C}\Dat E(t) &= \beta \, \frac{S(t) I(t)}{N_L(t)} - \sigma E(t), \\
{ }^{\C}\Dat I(t) &= \sigma E(t) - (\gamma + \mu) I(t), \\
{ }^{\C}\Dat R(t) &= \gamma I(t), \\
{ }^{\C}\Dat D(t) &= \mu I(t),
\end{aligned}\right.
\end{equation}
supplied with the initial conditions
\begin{equation}\label{E3.2}
S(0) = S_0 \ge 0, \quad E(0) = E_0 \ge 0, \quad I(0) = I_0 \ge 0, \quad R(0) = R_0 \ge 0, \quad D(0) = D_0 \ge 0.
\end{equation}
Here, $N_L(t) = S(t) + E(t) + I(t) + R(t)$ denotes the total living population, and $N = N_L(t) + D(t) = S(t) + E(t) + I(t) + R(t) + D(t)$ represents the total population. 
In the absence of natural births and deaths, $N$ remains constant over time, as shown in Section~\ref{S3.2}.

This fractional SEIRD formulation~\eqref{E3.1} generalizes the classical integer-order model: 
when $\alpha = 1$ (i.e. $\alpha \to 1^-$), the system reduces to the standard SEIRD model; 
when $\alpha < 1$, the system captures memory-driven dynamics, reflecting the influence of past states on present transmission. 
This property is particularly relevant in real epidemics, where delays in infection, incubation, and recovery processes often exhibit non-exponential waiting time distributions that cannot be adequately captured by ODEs.

\begin{remark}\label{R2}
In this formulation~\eqref{E3.1}, we do not include vital dynamics such as natural births and deaths. This choice is justified by the fact that epidemic outbreaks typically evolve on a much shorter timescale than demographic changes. 
By excluding vital dynamics, we ensure that the total population remains constant, allowing us to isolate the effects of disease-induced transitions and the fractional memory order $\alpha$, which is the main focus of this study.
\end{remark}

Before we proceed to the methodological aspects of model learning with PINNs, we must first establish the mathematical soundness of the proposed system. The following subsection examines the well-posedness of the fractional SEIRD model~\eqref{E3.1}--\eqref{E3.2}, focusing on the existence, uniqueness, and positivity of solutions under biologically meaningful conditions.

\subsection{Well-posedness Analysis}\label{S3.2}

This section establishes that the model~\eqref{E3.1}–\eqref{E3.2}is mathematically and biologically well-posed. 
Consider each equation in \eqref{E3.1} on the boundary of the non-negative orthant
\begin{equation*}
{ }^{\C}\Dat S(t)\big|_{S=0} = 0, \ { }^{\C}\Dat E(t)\big|_{E=0} = \beta \frac{S(t) I(t)}{N_L(t)} \ge 0, \ { }^{\C}\Dat I(t)\big|_{I=0} = \sigma E(t) \ge 0, 
\end{equation*}
\begin{equation*}
   { }^{\C}\Dat R(t)\big|_{R=0} = \gamma I(t) \ge 0, \ { }^{\C}\Dat D(t)\big|_{D=0} = \mu I(t) \ge 0.
\end{equation*}
Thus, on each coordinate hyperplane, the corresponding derivative points inward to $\mathbb{R}^5_{+}$. By the generalized mean value theorem for Caputo derivatives \cite[p. 288]{Odibat2007}, the solution remains non-negative for all $t \ge 0$.

Now, summing the first four equations of \eqref{E3.1} yields
\begin{equation*}
   { }^{\C}\Dat N_L(t) = - \mu I(t) \le 0.
\end{equation*}
Hence, $N_L$ is non-increasing and bounded above by its initial value, i.e.\ $N_L(0) = N_{L_0}$. Moreover,
\begin{equation*}
   { }^{\C}\Dat N(t) = 0,
\end{equation*}
which means $N(t)\equiv N(0) = N_0$. 
Therefore, every trajectory of \eqref{E3.1} remains in the compact positively invariant set
\begin{equation*}
     \Omega = \Big\{ (S,E,I,R,D)\in \mathbb{R}^5_{+} \, : \, S + E + I + R + D = N_0 \Big\}.
\end{equation*}

Let us denote $X(t) = \big(S(t), E(t), I(t), R(t), D(t)\big)^T$. 
The system~\eqref{E3.1} can be written in a compact form as
\begin{equation*}
     { }^{\C}\Dat X(t) = F(X(t)), \quad\text{where} \quad X(0) = X_0 \in \mathbb{R}^5_+,
\end{equation*}
where $F\colon\mathbb{R}^5_{+}\to \mathbb{R}^5$, the right-hand side of \eqref{E3.1}, is continuous and locally Lipschitz, because of the boundedness of $X$. By the general theory of fractional differential equations (see e.g. \cite{Diethelm2010}), the system admits a unique solution defined on $[0,\infty)$. The proposition below summarizes this section.

\begin{proposition}\label{P1}
For any given positive initial conditions and positive parameters $\beta$, $\sigma$, $\gamma$, $\mu$, the fractional system~\eqref{E3.1}–\eqref{E3.2} has a unique global solution $X(t)$ on $[0,\infty)$. 
Furthermore, this solution remains positive and bounded for all $t \geq 0$.
\end{proposition}

\subsection{Normalization and Scaling}\label{S3.3}

To facilitate the analysis and design of the PINN framework, we introduce dimensionless variables that represent proportions of the total population
\begin{equation*}
    s(t) = \frac{S(t)}{N}, \ e(t) = \frac{E(t)}{N}, \ i(t) = \frac{I(t)}{N}, \ r(t) = \frac{R(t)}{N}, \ \text{ and } \ d(t) = \frac{D(t)}{N},
\end{equation*}
   where $N_L(t)=N(1-d(t))$.
Dividing the fractional SEIRD model~\eqref{E3.1} by $N$, and using the linearity of the Caputo operator ${}^{\C}\Dat$, we obtain the normalized system
\begin{equation}\label{E3.3}
\left\{\begin{aligned}
{}^{\C}\Dat s(t) &= - \beta \frac{s(t)\, i(t)}{1 - d(t)}, \\
{}^{\C}\Dat e(t) &= \beta \frac{s(t)\, i(t)}{1 - d(t)} - \sigma e(t), \\
{}^{\C}\Dat i(t) &= \sigma e(t) - (\gamma + \mu) i(t), \\
{}^{\C}\Dat r(t) &= \gamma i(t), \\
{}^{\C}\Dat d(t) &= \mu i(t),
\end{aligned}\right.
\end{equation}
supplied with initial conditions
\begin{equation}\label{E3.4}
s(0) = s_0, \quad e(0) = e_0, \quad i(0) = i_0, \quad r(0) = r_0, \quad \text{and} \quad d(0) = d_0.
\end{equation}
Since ${}^{\C}\Dat (s+e+i+r+d)=0$, showing that the simplex
\begin{equation*}
\Delta = \bigl\{(s,e,i,r,d)\in \mathbb{R}^5_+ \ : \ s + e + i + r + d = 1 \text{ and } d < 1\bigr\},
\end{equation*}
is forward invariant. Thus, all state variables remain nonnegative and bounded by $1$ whenever the initial conditions belong to $\Delta$.

This normalized formulation~\eqref{E3.3}--\eqref{E3.4} offers several advantages for the subsequent PINN methodology:  
$(i)$ the state variables are naturally bounded in $[0,1]$, which improves numerical conditioning and facilitates the use of positivity or conservation constraints within the neural network;  
$(ii)$ working with proportions yields more easily interpretable and comparable parameters across different datasets;
$(iii)$ the explicit denominator $1-d(t)$ preserves the impact of mortality on transmission while keeping the equations scale-free.  
All methodological developments in the subsequent subsections are based on this normalized system~\eqref{E3.3}--\eqref{E3.4}.

\section{Methodology}\label{S4}

This section describes the proposed methodology for learning epidemic trajectories and estimating the parameters of the fractional SEIRD model. 
Our approach leverages PINNs, which integrate observational data with the governing fractional differential equations to ensure consistency with epidemic dynamics. 
Using system~\eqref{E3.3}--\eqref{E3.4}, the PINN simultaneously approximates the compartmental trajectories and recovers the parameters, including the fractional order $\alpha$, while enforcing positivity and conservation of the total population. 


\subsection{PINN Framework}\label{S4.1}

The key idea of PINNs is to approximate the solution to a system of differential equations using a neural network whose training is guided by both observational data and the governing equations. 
In our setting, the network learns the normalized epidemic trajectories
\begin{equation*}
    \bigl(s(t), e(t), i(t), r(t), d(t)\bigr) \in \Delta,
\end{equation*}
along with the unknown parameters $\beta$, $\sigma$, $\gamma$, $\mu$, and, most importantly, the fractional order $\alpha$.

\subsubsection{Network Architecture} 

We employ a fully connected feedforward neural network (multilayer perceptron) that takes time $t \in [0, T]$ as input and outputs approximations
\begin{equation*}
\mathcal{NN}^{\theta, \Theta} \colon t \mapsto \bigl(\hat{s}(t;\theta), \hat{e}(t;\theta), \hat{i}(t;\theta), \hat{r}(t;\theta), \hat{d}(t;\theta) \bigr),
\end{equation*}
where $\theta$ and $\Theta$ denote the trainable weights and biases of the network, respectively.

To maintain epidemiological validity, we enforce nonnegativity and approximate population conservation through the output layer. 
This can be achieved either by using  \emph{softplus}, an activation function defined as $\emph{softplus}(x) = \ln(1+e^x)$. 
This function smoothly maps real numbers to positive values, ensuring nonnegativity without sharp cutoffs, or activations to guarantee positivity. Alternatively, we can apply a \emph{softmax}, a normalization function that transforms a vector into positive components that sum to one: $\emph{softmax}(z_i) = \frac{e^{z_i}}{\sum_j e^{z_j}}$. 
It is commonly used to represent probabilities or normalized fractions. The softmax projection ensures
\begin{equation*}
   \hat{s}(t)+\hat{e}(t)+\hat{i}(t)+\hat{r}(t)+\hat{d}(t) \approx 1,
\end{equation*}
so that the total population is conserved. This constraint can be imposed explicitly via a simplex-based architecture or softly enforced through a penalty term in the loss function (see Sections~\ref{S4.4} and \ref{S4.5}).

\subsubsection{Physics-Informed Residuals}

Rather than training solely on observed data, the PINN incorporates the fractional SEIRD equations~\eqref{E3.3} into the loss function by ensuring that the neural outputs satisfy the dynamics in the sense of the Caputo derivative. 
Specifically, the fractional derivatives ${}^\C\Dat \hat{s}(t)$, ${}^\C\Dat \hat{e}(t)$, etc., are approximated numerically (see Section~\ref{S4.2}) and substituted into the right-hand side of the system. 
The resulting residuals vanish for the true solution; therefore, minimizing them enforces consistency between the neural predictions and the governing epidemic model.

\subsubsection{Joint Learning of Trajectories and Parameters}

The network weights $\theta$ and the epidemiological parameters $\Theta = \{\beta, \sigma, \gamma, \mu, \alpha \}$ are updated simultaneously. 
This hybrid approach allows the model to reconstruct unobserved states and estimate parameters from limited data. The embedded physics serve as a regularizer, mitigating overfitting and enhancing interpretability.

\subsection{Caputo Derivative Discretization}\label{S4.2}

For a sufficiently smooth function $\psi$, we recall the Caputo derivative ${}^\C\Dat \psi$ defined in \eqref{E2.1}.
Its nonlocal nature introduces long-range memory effects, which are central to our analysis. 
To efficiently approximate the derivative at discrete time points $t_n = n \Delta t$, we adopt the classical L1 scheme (see, e.g., \cite[p. 34-40]{Gao2014})
\begin{equation}\label{E4.1}
{}^\C\Dat \psi(t_n) \;\approx\; \frac{1}{\Delta t^{\alpha}} 
\sum_{k=0}^{n-1} c_{k}^{(n)} \, \bigl( \psi(t_{n-k}) - \psi(t_{n-k-1}) \bigr),
\end{equation}
with weights
\begin{equation*}
c_k^{(n)} = \frac{1}{\Gamma(2-\alpha)} \bigl[ (k+1)^{1-\alpha} - k^{1-\alpha} \bigr], \ \text{ with } \ k=0,1,\dots,n-1.
\end{equation*}
This discretization plays a crucial role in the PINN framework developed in this paper because it allows fractional-order dynamics to be directly embedded into the loss function. 
The L1 scheme~\eqref{E4.1} is differentiable with respect to both $\psi(t)$ and $\alpha$, enabling gradients to propagate through the discretization during training. 
In practice, the L1 operator is implemented as a differentiable layer within the PINN to ensure efficient backpropagation.

The neural network outputs $\hat{s}(t)$, $\hat{e}(t)$, $\hat{i}(t)$, $\hat{r}(t)$, and $\hat{d}(t)$ are evaluated at collocation points $\{t_j\}$, and their Caputo derivatives are computed using the L1 scheme~\eqref{E4.1}. 
These approximations are substituted into the residuals of the fractional SEIRD system to ensure that the network predictions satisfy the underlying dynamics.

\subsection{Parameter Estimation Strategy}\label{S4.3}

The PINN is designed not only to approximate epidemic trajectories but also to identify the epidemiological parameters
\begin{equation*}
   \Theta = \{\beta, \sigma, \gamma, \mu, \alpha\}.
\end{equation*}
All parameters are treated as trainable variables and updated jointly with the network weights. To ensure admissibility and stability,  the following design choices are imposed:
\begin{itemize}
\item[-] \textbf{Positivity:} All rates are parameterized via \emph{softplus} transformations, e.g.\ $\beta = \emph{softplus}(\mathscr{C}_\beta)$, ensuring $\beta > 0$, where $\mathscr{C}_\beta \in \mathbb{R}$ and $\beta \le \beta_{max}$.

\item[-] \textbf{Fractional Order:} The memory parameter $\alpha$ is restricted to $(\alpha_{\min},1]$ using a sigmoid mapping
\begin{equation*}
    \alpha = \alpha_{\min} + (1-\alpha_{\min})\, \sigma(z_\alpha),
\end{equation*}
where $z_\alpha \in \mathbb{R}$ and $\sigma(\cdot)$ denotes the sigmoid function.

\item[-] \textbf{Staged Optimization:} Training proceeds in two phases:
\begin{itemize}
\item[$(i)$] an initialization phase where $\alpha$ is fixed at 1 (classical SEIRD) and only the epidemiological parameters are trained;

\item[$(ii)$] a joint phase where $\alpha$ is released and estimated simultaneously with the other parameters.
\end{itemize}
\end{itemize}
Initial guesses for $\Theta$ are obtained from classical SEIRD fits or epidemiological literature, and parameter bounds are imposed to exclude unrealistic values.

\subsection{Loss Function}\label{S4.4}

In the PINN framework, the loss function plays a central role in guiding the neural network to approximate both the epidemic trajectories and the underlying dynamics of the fractional SEIRD system~\eqref{E3.3}--\eqref{E3.4}. 
Unlike purely data-driven models, which only minimize prediction error, here the loss function in the PINN framework must balance multiple objectives: fitting available observations, enforcing the governing fractional differential equations, satisfying initial conditions, and maintaining epidemiological consistency, such as positivity and population conservation. 
To achieve this balance, we construct a composite loss function that integrates data misfit with physics-based residuals and additional structural constraints. 
This design ensures that the trained network accurately reflects the observed data while remaining faithful to the model's mathematical and biological principles.

The PINN is trained by minimizing the composite loss functional
\begin{equation}\label{E4.2} 
\mathcal{L}(\theta,\Theta) = \lambda_{\text{data}}\mathcal{L}_{\text{data}}
+ \lambda_{\text{phys}}\mathcal{L}_{\text{phys}}
+ \lambda_{\text{IC}}\mathcal{L}_{\text{IC}}
+ \lambda_{\text{cons}}\mathcal{L}_{\text{cons}}
+ \lambda_{\text{reg}}\mathcal{L}_{\text{reg}}.
\end{equation}
Each component is defined as follows:
\begin{itemize}
\item[] \textbf{Data Misfit:} Penalizes discrepancies between predicted and observed compartments $\mathcal{O} = \{s,e,i,r,d\}$,
\begin{equation*}
   \mathcal{L}_{\text{data}} = \frac{1}{N_{\text{data}}} \sum_{j = 1}^{N_{\text{data}}} \sum_{\mathrm{x} \in \mathcal{O}} \big|\hat{\mathrm{x}}(t_j)-\mathrm{x}_j^{\text{obs}}\big|^2.
\end{equation*}

\item[] \textbf{Physics Residual:} Enforces consistency with the fractional SEIRD dynamics at collocation points
\begin{equation*}
   \mathcal{L}_{\text{phys}} = \frac{1}{N_{\text{coll}}}\sum_{j=1}^{N_{\text{coll}}}\sum_{\mathrm{x} \in \mathcal{O}} \big| \mathcal{R}_\mathrm{x}(t_j) \big|^2,
\end{equation*}
where $\mathcal{R}_\mathrm{x}(t) = { }^\C\Dat\hat{\mathrm{x}}(t) - F_\mathrm{x}(\hat{\mathrm{x}},\beta,\sigma,\gamma,\mu)$ is the residual of the governing equation~\eqref{E3.3}, and ${ }^\C\Dat$ is discretized using the L1 scheme~\eqref{E4.1}.

\item[] \textbf{Initial Conditions:} Ensures agreement with prescribed initial values
\begin{equation*}
\mathcal{L}_{\text{IC}} = \sum_{\mathrm{x}\in \mathcal{O}} \big|\hat{\mathrm{x}}(0)-\mathrm{x}_0\big|^2.
\end{equation*}

\item[] \textbf{Conservation:} Penalizes violations of the population balance $\hat{s}+\hat{e}+\hat{i}+\hat{r}+\hat{d}=1$ at collocation points
\begin{equation*}
    \mathcal{L}_{\text{cons}} = \frac{1}{N_{\text{coll}}}\sum_{j=1}^{N_{\text{coll}}}\Big(\hat{s}(t_j)+\hat{e}(t_j)+\hat{i}(t_j)+\hat{r}(t_j)+\hat{d}(t_j)-1\Big)^2.
\end{equation*}

\item[] \textbf{Regularization:} $\ell^2$ penalties on the network weights (and optionally the parameters $\theta$) to avoid overfitting
\begin{equation*}
    \mathcal{L}_{\text{reg}} = \lambda_\Theta \|\Theta\|_2^2 + \lambda_\theta \|\theta\|_2^2.
\end{equation*}
\end{itemize}
The weights $\lambda_*$ balance the contributions of each term and are tuned empirically to achieve stability between data fidelity and physics enforcement.

\begin{figure}[H]
\centering
\begin{tikzpicture}
\draw[rounded corners=10pt, dashed, thick, blue] (-4.8, -1) rectangle (3.5, 3) {};
\node[circle, draw, fill=blue!30, minimum size=0.9cm, thick, xshift=-1cm, yshift=1cm] (input) at (-3, 0) {$t$};
\node[circle, draw, fill=gray!30, minimum size=0.5cm, thick, yshift=1cm] (h1) at (-2, 1.5) {};
\node[circle, draw, fill=gray!30, minimum size=0.5cm, thick, yshift=1cm] (h2) at (-2, 0.5) {};
\node[circle, draw, fill=gray!30, minimum size=0.5cm, thick, yshift=1cm] (h3) at (-2, -0.5) {};
\node[circle, draw, fill=gray!30, minimum size=0.5cm, thick, yshift=1cm] (h4) at (-2, -1.5) {};
\node[circle, draw, fill=gray!30, minimum size=0.5cm, thick, yshift=1cm] (h5) at (-0.5, 1.5) {};
\node[circle, draw, fill=gray!30, minimum size=0.5cm, thick, yshift=1cm] (h6) at (-0.5, 0.5) {};
\node[circle, draw, fill=gray!30, minimum size=0.5cm, thick, yshift=1cm] (h7) at (-0.5, -0.5) {};
\node[circle, draw, fill=gray!30, minimum size=0.5cm, thick, yshift=1cm] (h8) at (-0.5, -1.5) {};
\node[rectangle, rounded corners, draw, fill=green!30, minimum size=0.9cm, thick, xshift=0.5cm, yshift=1cm] (output) at (1.5, 0) { \ $\hat{s},\hat{e},\hat{i},\hat{r},\hat{d}$ \ };
\foreach \i in {h1,h2,h3,h4}
  \draw[->, thick] (input) -- (\i);
\foreach \i in {h5,h6,h7,h8}
  \foreach \j in {h1,h2,h3,h4}
    \draw[->, thick] (\j) -- (\i);
\foreach \i in {h5,h6,h7,h8}
  \draw[->, thick] (\i) -- (output);
\node[anchor=north,above] at (-0.9, 3.1) {\color{blue!70!black}Neural Network: $\mathcal{NN}^{\theta,\Theta}(t)$};
\draw[rounded corners=10pt, dashed, orange, thick, yshift=0.9cm] (0.2, -1) rectangle (11.5, 1.2) {};
\node[circle, draw, fill=orange!30, thick, yshift=1cm] (dudt) at (4.7, 0) {${}^\C\Dat$};
\node[rectangle, rounded corners, draw, fill=yellow!30, thick, align=center, yshift=1cm, text width=4.3cm] (residual) at (8.5, 0) {$\mathcal{R}_\mathrm{x}(t) = {}^\C\Dat\hat{\mathrm{x}}(t) - F_\mathrm{x}(\hat{\mathrm{x}},\beta,\sigma,\gamma,\mu)$};
\draw[->, thick] (output) -- (dudt);
\draw[->, thick] (dudt) -- (residual);
\node[anchor=north, above, yshift=1cm] at (6.25, 1.25) { \quad \color{orange!70!black}Fractional SEIRD System~\eqref{E3.3}};
\node[rectangle, rounded corners, draw, fill=cyan!20, thick, align=center, xshift=-0.5cm, yshift=0.8cm] (loss) at (4, -2.5) { \ LOSS: $\mathcal{L} = \mathcal{L}_{\text{data}} + \mathcal{L}_{\text{phys}} + \mathcal{L}_{\text{IC}} + \mathcal{L}_{\text{cons}} + \mathcal{L}_{\text{reg}}$ \ };
\draw [->, thick] (output.south) -| (2.5,-1.35);
\draw [->, thick] (residual.south) -| (8.5,0) -- (8.5,-1.7) -- (loss.east);
\end{tikzpicture}
\captionof{figure}{Schematic diagram of the proposed PINN for the fractional SEIRD model~\eqref{E3.3}--\eqref{E3.4}.}\label{F1}
\end{figure}
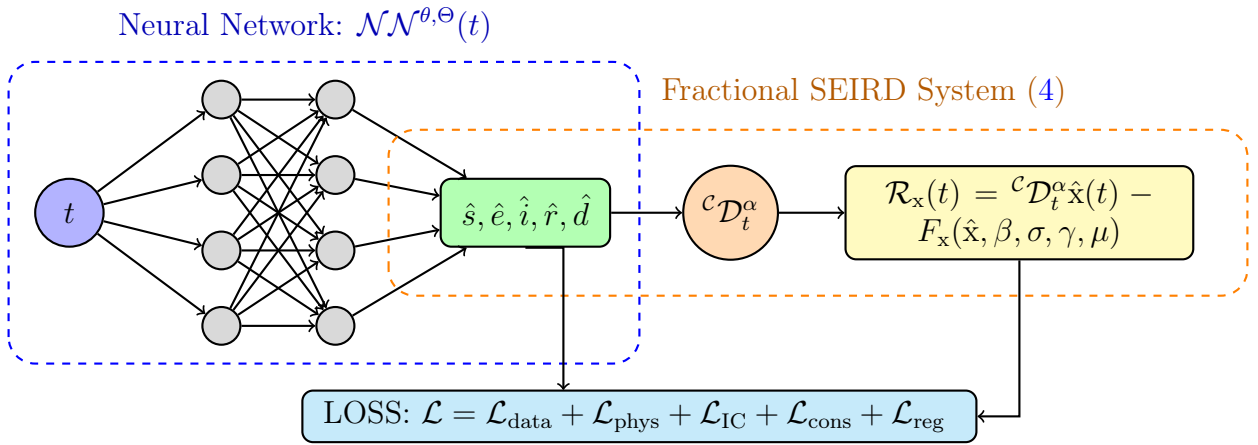

Figure~\ref{F1} summarizes the complete architecture of the proposed PINN. The neural network receives time $t$ as input and outputs approximations of the compartmental trajectories $(\hat{s},\hat{e},\hat{i},\hat{r},\hat{d})$. 
These outputs are processed through the Caputo derivative operator ${}^\C\Dat$, discretized using the L1 scheme~\eqref{E4.1}, and compared against the fractional SEIRD dynamics~\eqref{E3.3} to form physics residuals. 
Along with the data misfit, physics and initial condition enforcement, conservation penalties, and regularization, these residuals contribute to the composite loss~\eqref{E4.2}, which guides the training.

\subsection{Training Algorithm}\label{S4.5}

The training procedure integrates the loss function defined in Section~\ref{S4.4} with staged optimization strategies that are designed to improve the stability and identifiability of the fractional order $\alpha$. The workflow is summarized below.
\begin{algorithm}[H]
\caption{Fractional-SEIRD PINN Training Algorithm}\label{Alg1}
\begin{algorithmic}[1]
\STATE \emph{Initialization:} 
Neural network weights $\Theta$ are initialized using Xavier initialization \cite{Glorot2010}. Epidemiological parameters $\theta=(\beta,\sigma,\gamma,\mu)$ are drawn from admissible ranges informed by prior studies, while the fractional order $\alpha$ is initialized close to $1$ to reflect the classical Markovian case (see Remark~\ref{remfoot3})

\STATE \emph{Staged Optimization:} 
Training proceeds in two phases:
\vspace{-0.3cm}
\begin{itemize}
    \item[$i:$] \emph{Pretraining}: Fix $\alpha=1$ and minimize a reduced loss $\mathcal{L}_{\text{data}}+\mathcal{L}_{\text{IC}}$ to fit observations and initial conditions;
    \vspace{-0.3cm}
    \item[$ii:$] \emph{Joint training}: Release $\alpha$ and optimize the full composite loss $\mathcal{L}$ in \eqref{E4.2}, updating both network weights and epidemiological parameters.  
    \vspace{-0.3cm}
\end{itemize} 

\STATE \emph{Optimization Scheme:}  
  We employ a hybrid optimizer \cite{Smith2017, Liu1989, Raissi2019}: Adam with a decaying learning rate is used for initial exploration, followed by L-BFGS for fine-tuning. This hybrid strategy has been shown to be effective in PINNs.

\STATE \emph{Constraints and Monitoring:}  
  Positivity and conservation are enforced through the loss penalties described in Section~\ref{S4.4}. Training is monitored via the total loss and its components. Early stopping is applied when improvements fall below the tolerance threshold.
\end{algorithmic}
\end{algorithm}
\begin{remark}\label{remfoot3}
When $\alpha=1$, the fractional derivative reduces to the classical first-order derivative, resulting in the standard Markovian SEIRD model with memoryless dynamics. Therefore, initializing $\alpha$ near 1 therefore, lets the model begin training close to the classical case and adapt toward sub-exponential, memory-dependent dynamics if the data supports it.
\end{remark}


Algorithm~\ref{Alg1} simultaneously reconstructs latent epidemic trajectories and reliably estimates the fractional order $\alpha$, thereby quantifying memory effects while maintaining epidemiological interpretability.

\section{Results and Discussion}\label{S5}

To evaluate the proposed fractional SEIRD model~\eqref{E3.3}--\eqref{E3.4} with PINNs, we conduct experiments on both synthetic and real epidemic datasets, designed to address the following questions:
\begin{itemize}
\item[-] Can the framework accurately recover the fractional memory order $\alpha$ and epidemiological parameters from noisy data?

\item[-] How does the inclusion of memory effects improve predictive performance compared to classical integer-order SEIRD models?

\item[-] What is the role of different architectural and methodological choices (e.g., fixed vs trainable $\alpha$, conservation constraints) in the model's performance?

\item[-] How reliable are the estimated parameters and predictions, given uncertainty and data limitations?
\end{itemize}

To this end, we first validate the approach using synthetic epidemic trajectories that mimic the dynamics of the Mpox epidemic, for which the ground-truth parameters $(\beta, \sigma, \gamma, \mu, \alpha)$ are known. 
We simulate the dynamics of the fractional SEIRD model~\eqref{E3.3} through the L1 discretization of the Caputo derivative~\eqref{E4.1} for various values of $\alpha \in (0,1]$, with Gaussian noise added to replicate observation errors. 
Next, we use the procedure in Section~\ref{S4.5} to train the PINN to infer $\alpha$ and the epidemiological parameters from partial and noisy observations. This confirms the identifiability and stability of the proposed learning strategy before turning to real-world data. 
We then apply the method to COVID-19 datasets, focusing on confirmed, recovered, and deceased case counts from several European regions. We assess performance through trajectory fitting, estimation of the optimal fractional order $\alpha$, and comparison with the predictive accuracy of the fractional SEIRD model. 

Section~\ref{S5.1} presents the results for the synthetic setting, based on numerical solutions obtained using the L1 discretization scheme~\eqref{E4.1}. Section~\ref{S5.2} details the evaluation on real datasets, emphasizing predictive accuracy, parameter estimation, and epidemiological interpretation.

\subsection{Synthetic Data Experiments}\label{S5.1}

The parameters were selected within biologically plausible ranges reported in the literature for Mpox, as shown in Table~\ref{Tab1}.
\begin{table}[H]
\centering
\setlength{\tabcolsep}{0.5cm}
\caption{Epidemiological parameters for the SEIRD model applied to Mpox (clade Ib, 2024–2025).}\label{Tab1}
\adjustbox{max width=\textwidth}{
\begin{tabular}{cccccc}
\hline
\textbf{Epidemic} & $\beta$ & $\sigma$ & $\gamma$ & $\mu$ & \textbf{References} \\
\hline\hline
Mpox & 0.1 – 0.3 & 0.077 – 0.2 & 0.036 – 0.071 & 0.001 – 0.03 & \cite{WHO2024Mpox, Lancet2024} \\
\hline
\end{tabular}
}
\end{table}

Specifically, the fixed values were set to
\begin{equation*}
    \beta = 0.25, \quad \sigma = 0.13, \quad \gamma = 0.052, \quad \mu = 0.005,
\end{equation*}
while the memory order varied as $\alpha \in \{1, 0.95, 0.9\}$. 
The case $\alpha = 1$ corresponds to the classical integer-order SEIRD model, while $\alpha < 1$ incorporates fractional memory effects.
The values $\alpha = 0.95$ and $\alpha = 0.9$ were chosen to introduce moderate deviations from the classical case. These values remain close enough to $\alpha = 1$ to ensure biological plausibility and numerical stability, while still reflecting realistic long-memory effects, such as incubation variability, reporting delays, or behavioral adaptation.

\begin{figure}[H]
\centering
\begin{subfigure}{0.9\textwidth}
\includegraphics[width=\linewidth]{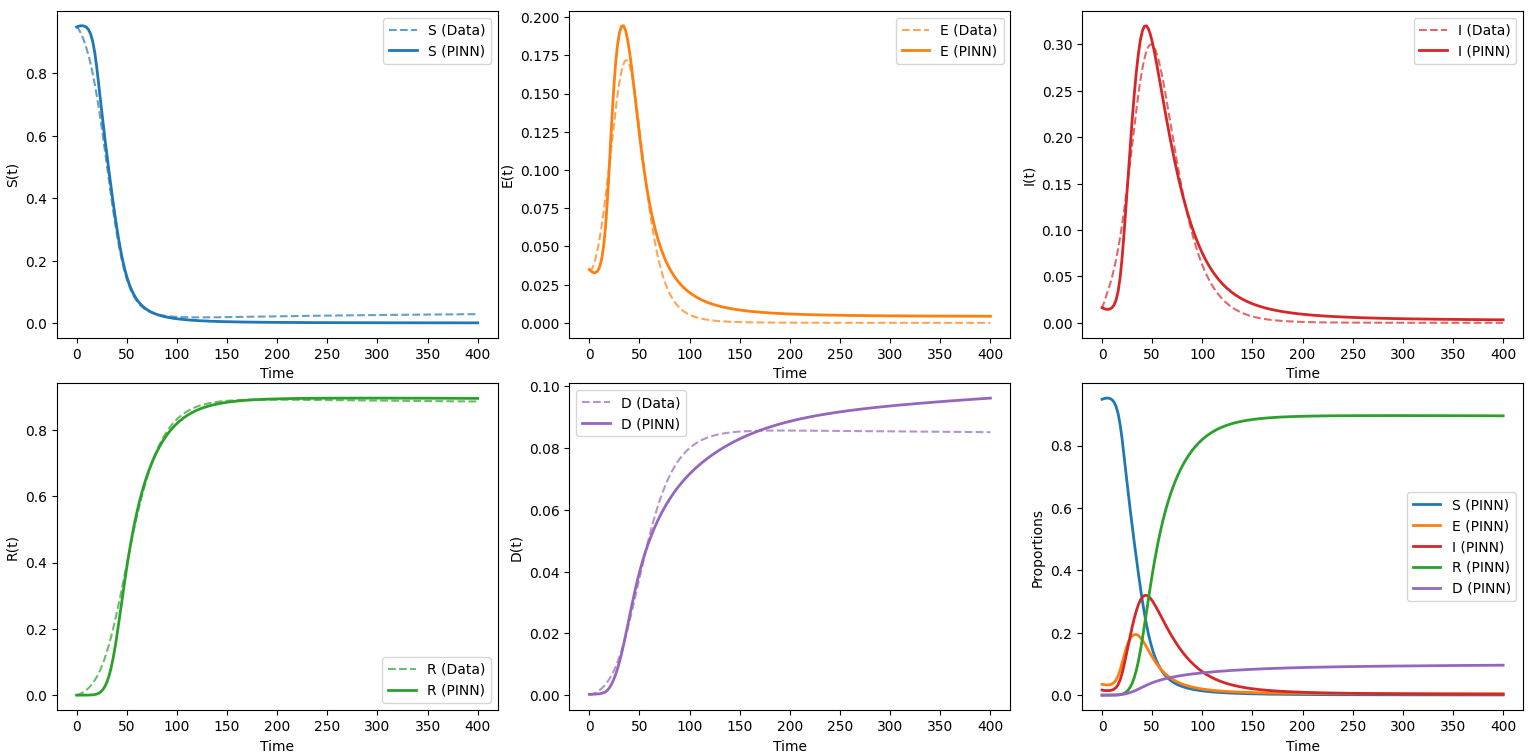}
\caption{$\alpha = 1$}
\end{subfigure}
\hfill 
\begin{subfigure}{0.9\textwidth}
\includegraphics[width=\linewidth]{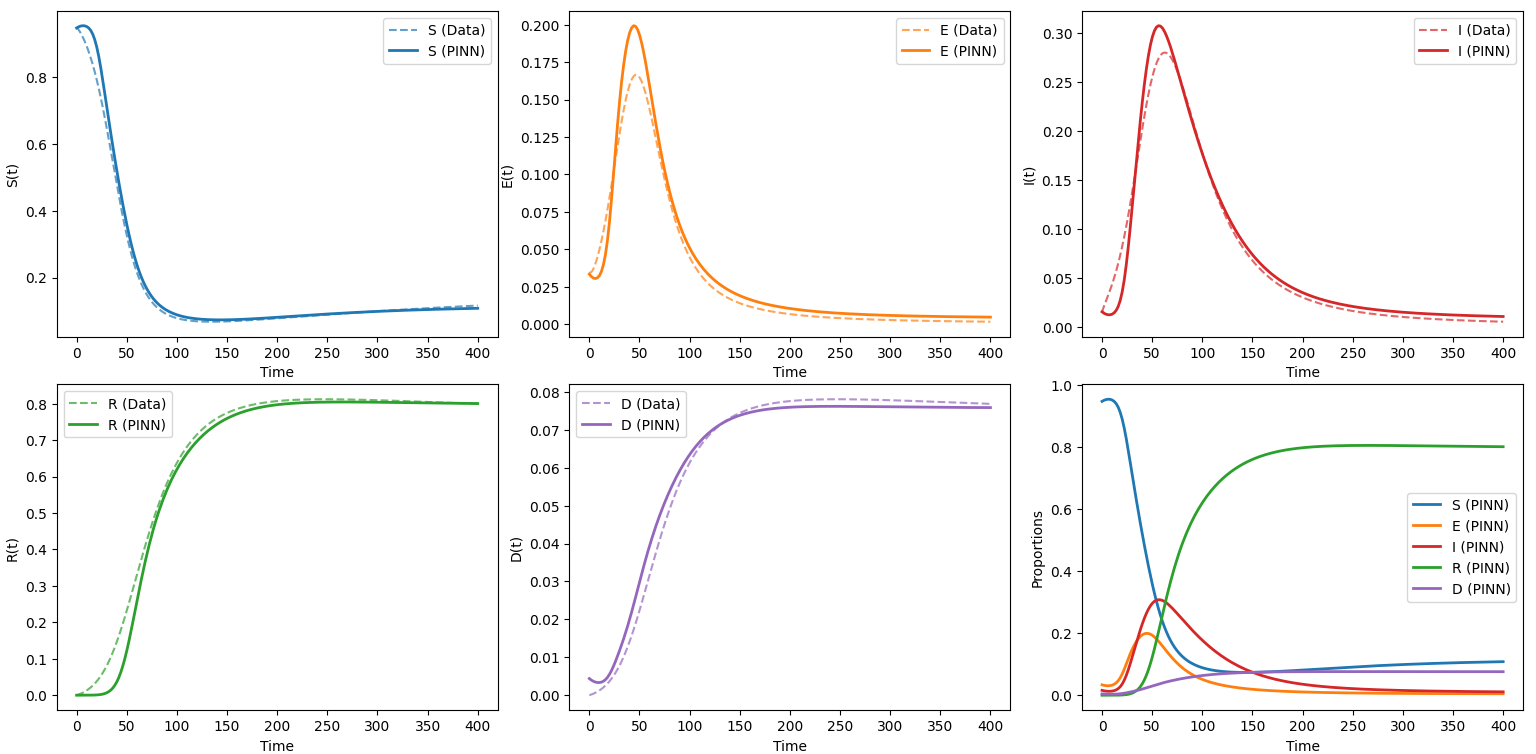}
\caption{$\alpha = 0.95$}
\end{subfigure}
\hfill 
\begin{subfigure}{0.9\textwidth}
\includegraphics[width=\linewidth]{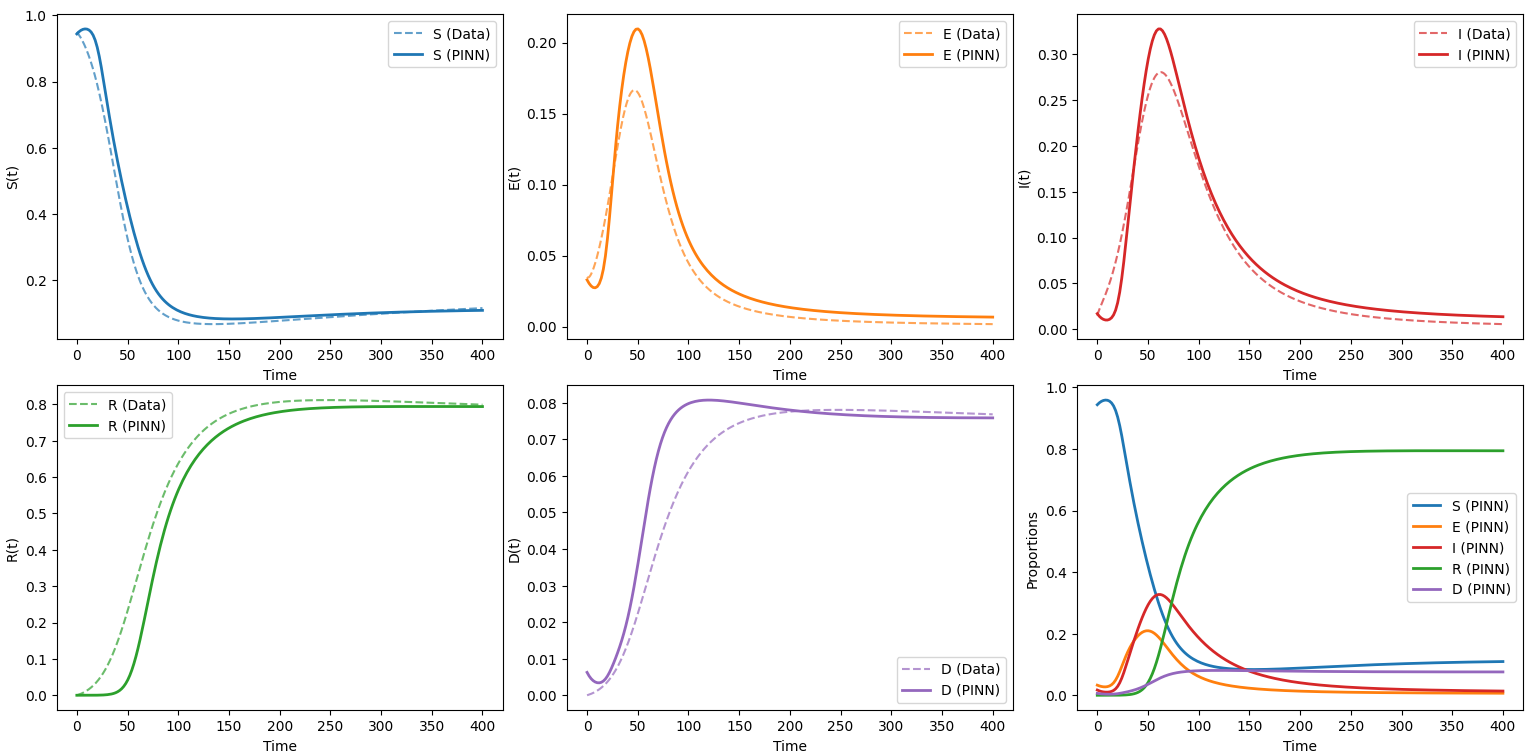}
\caption{$\alpha = 0.9$}
\end{subfigure}
\caption{PINN predictions of SEIRD dynamics under different fixed values of $\alpha$.}\label{F2}
\end{figure}

\begin{figure}[H]
\centering
\begin{subfigure}{0.9\textwidth}
\includegraphics[width=\linewidth]{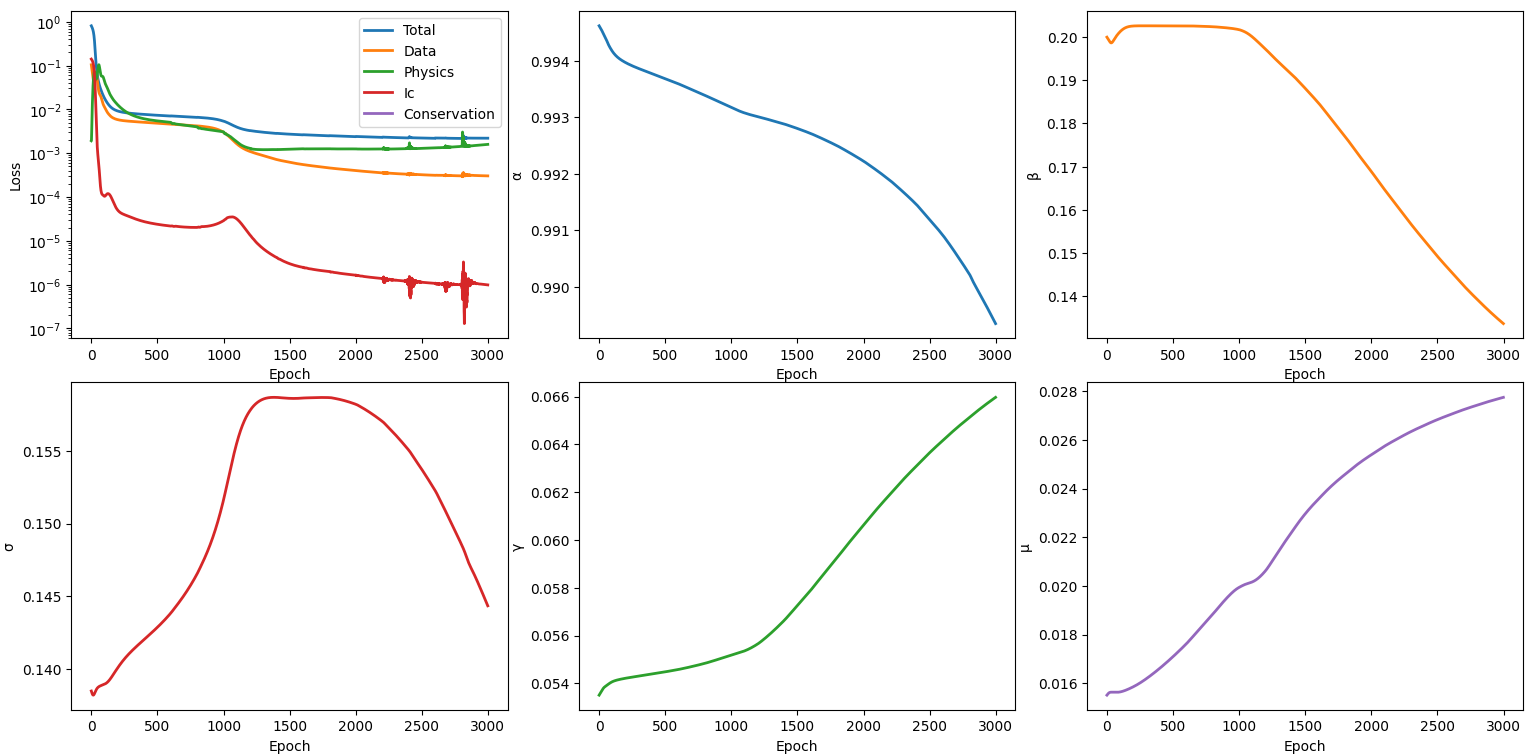}
\caption{$\alpha = 1$}
\end{subfigure}
\hfill 
\begin{subfigure}{0.9\textwidth}
\includegraphics[width=\linewidth]{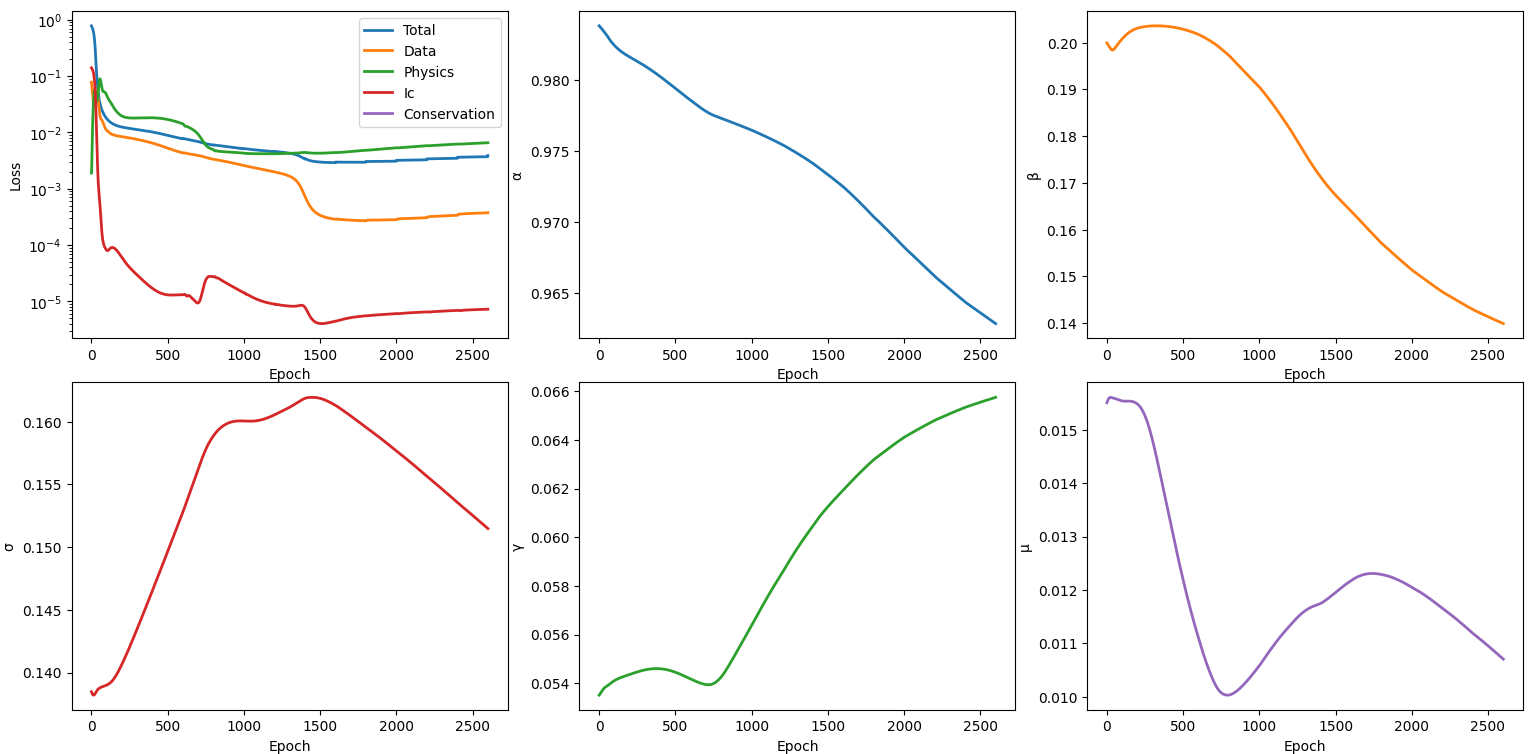}
\caption{$\alpha = 0.95$}
\end{subfigure}
\hfill 
\begin{subfigure}{0.9\textwidth}
\includegraphics[width=\linewidth]{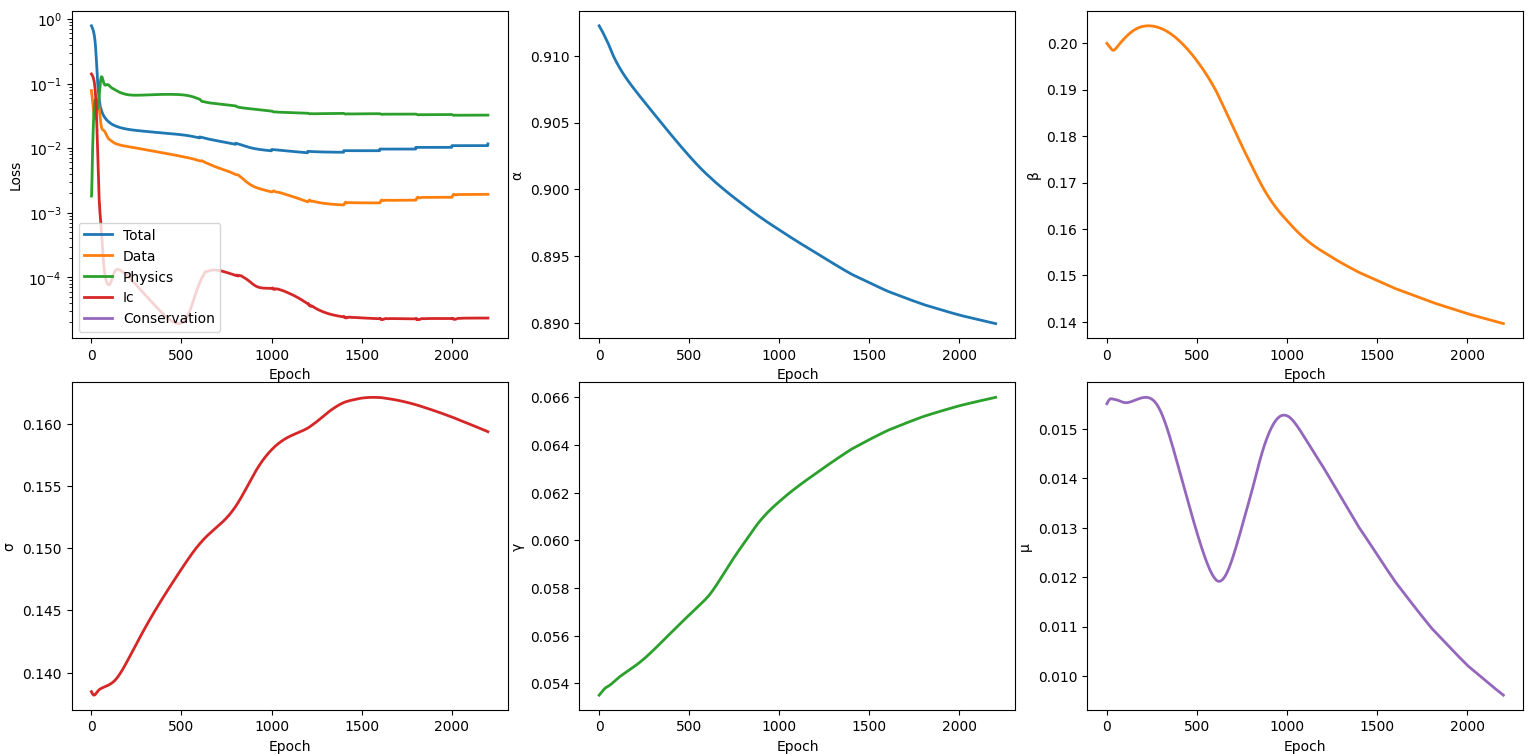}
\caption{$\alpha = 0.9$}
\end{subfigure}
\caption{Estimated epidemiological parameters under different fixed $\alpha$.}\label{F3}
\end{figure}

Figures~\ref{F2} and~\ref{F3} provide a visual comparison of the reconstructed epidemic dynamics and the corresponding parameter estimates under different fixed values of $\alpha$. 
Figure~\ref{F2} shows that the predicted trajectories for all compartments ($S, E, I, R, D$) closely align with the ground-truth synthetic data, demonstrating the PINN's  ability to capture both the transient and steady-state behaviors of the system. 
As $\alpha$ decreases, the infectious peak is slightly delayed and flattened, highlighting the moderating role of memory effects in epidemic progression.
Figure~\ref{F3} complements this analysis by illustrating the estimated epidemiological parameters across the three scenarios. 
Table~\ref{Tab2} summarizes these estimated values.  

The recovered parameters remain within the biologically relevant ranges listed in Table~\ref{Tab1}, demonstrating good consistency with the ground-truth settings. 
Notably, as $\alpha$ decreases, $\sigma$ increases slightly, while $\mu$ decreases. 
This trend aligns with the idea that fractional dynamics introduce memory effects that slow down epidemic progression while effectively reducing its severity. 
Meanwhile, $\beta$ and $\gamma$ remain relatively stable across all cases.
\begin{table}[H]
\centering
\setlength{\tabcolsep}{0.5cm}
\caption{Estimated epidemiological parameters for Mpox  with $\alpha$ fixed at 1, 0.95, and 0.9.}\label{Tab2}
\adjustbox{max width=\textwidth}{
\begin{tabular}{c||ccccc}
\hline
\textbf{Parameter} & $\alpha$ & $\beta$ & $\sigma$ & $\gamma$ & $\mu$ \\
\hline\hline
$\alpha = 1$ & 0.989349 & 0.133711 & 0.144335 & 0.065966 & 0.027749 \\
\hline
$\alpha = 0.95$ & 0.962859 & 0.139901 & 0.151477 & 0.065746 & 0.010705 \\
\hline
$\alpha = 0.9$ & 0.914439 & 0.136780 & 0.155774 & 0.066356 & 0.008520 \\
\hline
\end{tabular}
}
\end{table}

\subsection{Application to Real-World COVID-19 Data}\label{S5.2}

To further assess the applicability of the proposed framework further, we examined real-world epidemic data from Germany and Sweden during the COVID-19 outbreaks. 
We selected these two countries due to the availability of their daily case and mortality reports and their contrasting epidemic trajectories within Europe. 
The corresponding epidemiological parameters were constrained within biologically plausible ranges, as shown in Table~\ref{Tab3}.
\begin{table}[H]
\centering
\setlength{\tabcolsep}{0.5cm}
\caption{Epidemiological parameters for the SEIRD model applied to COVID-19 in Europe (2024).}\label{Tab3}
\adjustbox{max width=\textwidth}{
\begin{tabular}{cccccc}
\hline
\textbf{Epidemic} & $\beta$ & $\sigma$ & $\gamma$ & $\mu$ & \textbf{References} \\
\hline\hline
COVID-19 & 0.2 – 0.4 & 0.1 – 0.3 & 0.05 – 0.1 & 0.001 – 0.01 & \cite{UKGov2025, WHO2024Covid} \\
\hline
\end{tabular}
}
\end{table}

The daily number of confirmed cases and deaths were extracted for each country from publicly available databases \cite{OWD2025, Gehrcke2025, OWD2025Sweden, worldometers}. 
Cumulative counts were transformed into compartmental trajectories consistent with the SEIRD structure. 
Specifically, the infected class $I$ was reconstructed from active case estimates, the deceased class $D$ from reported daily deaths, and the removed class $R$ by aggregating recoveries and deaths, when recovery data were available. 
Since the exposed population $E$ was not directly observed, it was treated as a latent state to be inferred by the model.

\begin{figure}[H]
\centering
\begin{subfigure}{0.9\textwidth}
\includegraphics[width=\linewidth]{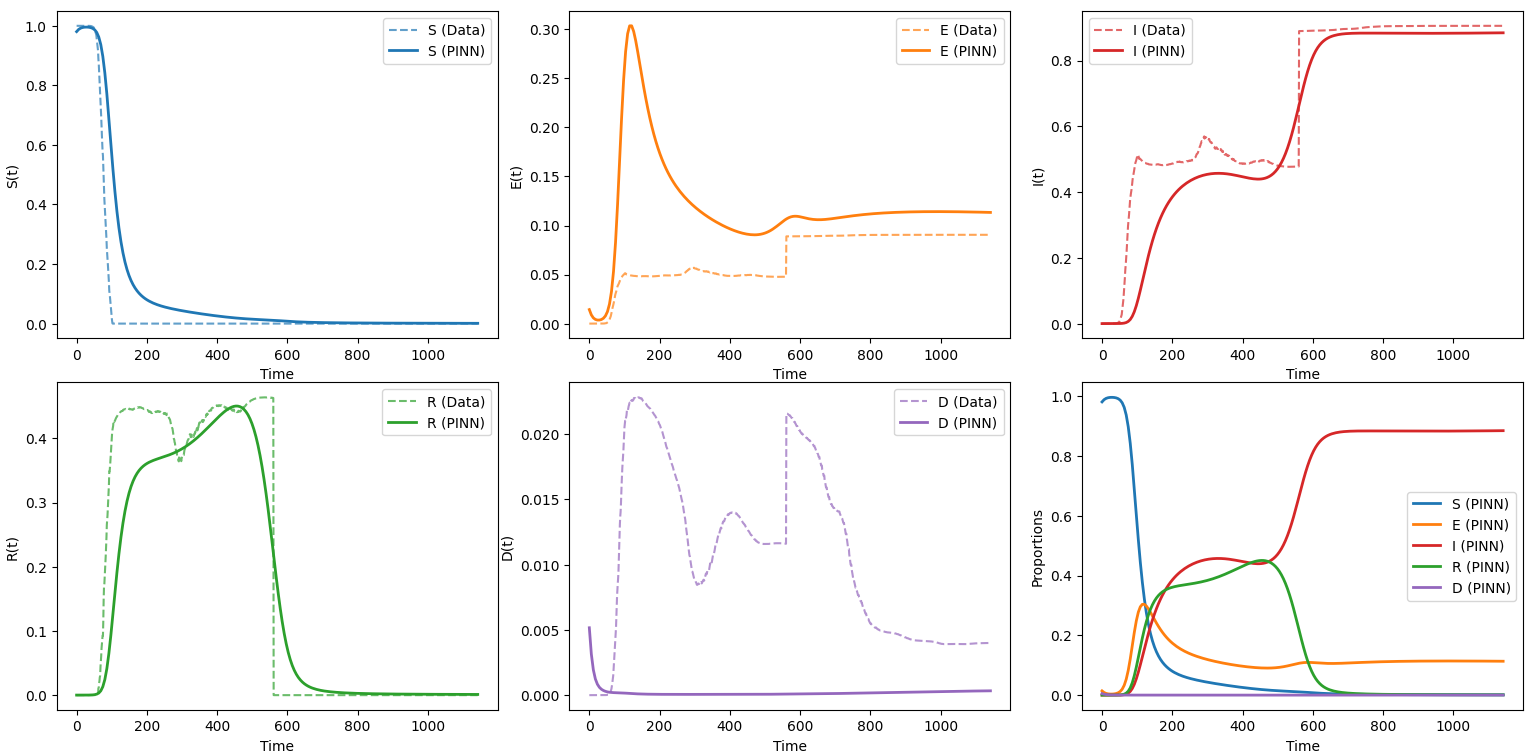}
\caption{Germany data}
\end{subfigure}
\hfill 
\begin{subfigure}{0.9\textwidth}
\includegraphics[width=\linewidth]{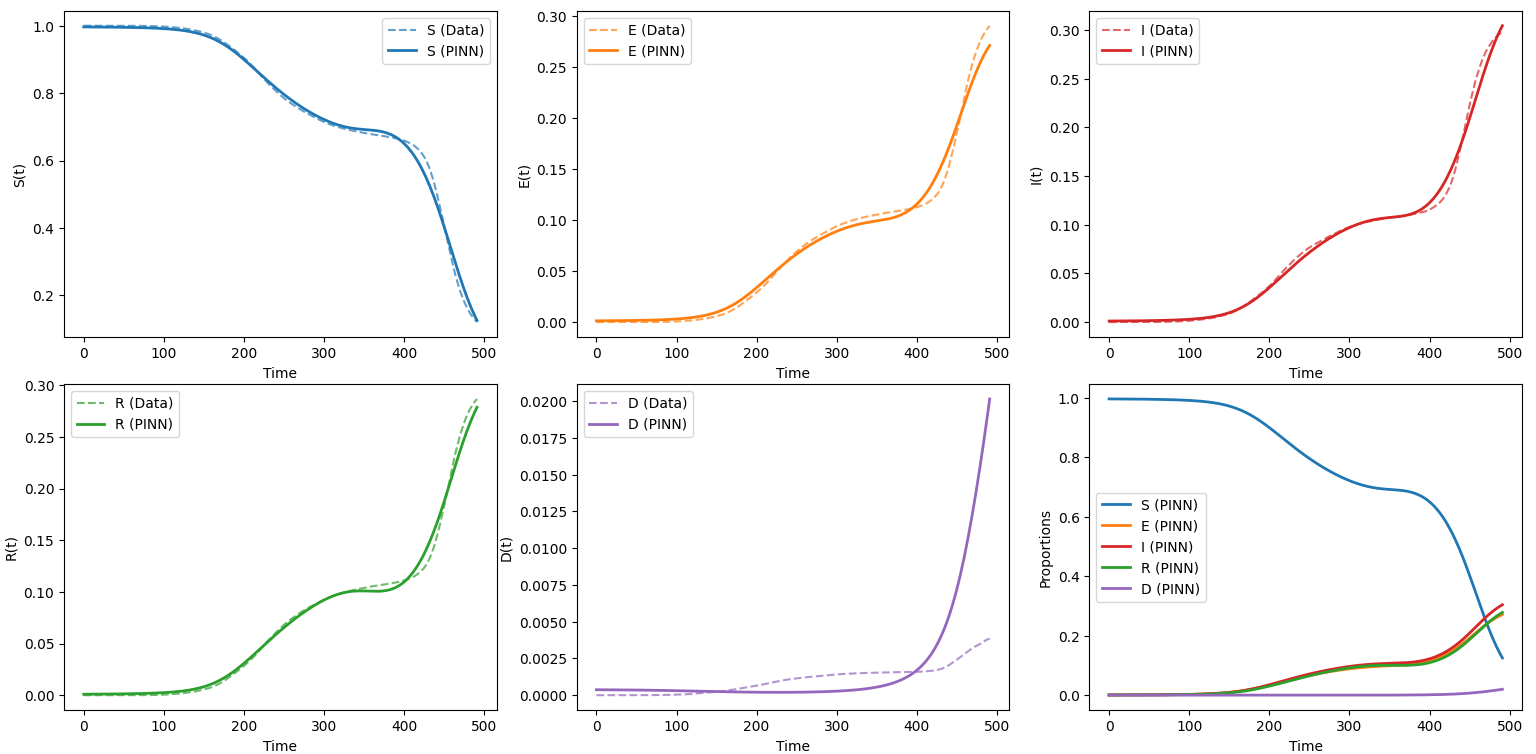}
\caption{Sweden data}
\end{subfigure}
\caption{PINN predictions of SEIRD dynamics for real data.}\label{F4}
\end{figure}

\begin{figure}[H]
\centering
\begin{subfigure}{0.9\textwidth}
\includegraphics[width=\linewidth]{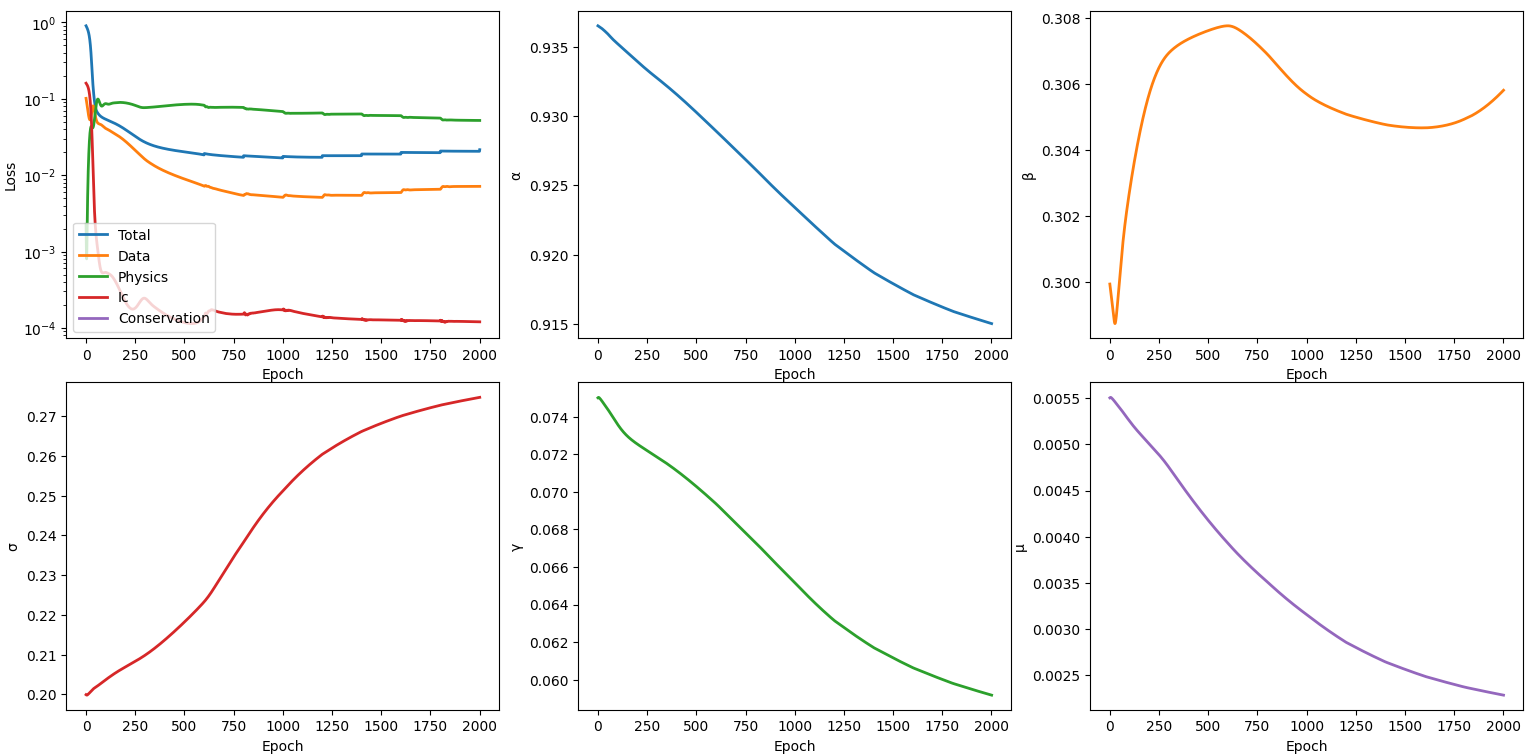}
\caption{Germany data}
\end{subfigure}
\hfill 
\begin{subfigure}{0.9\textwidth}
\includegraphics[width=\linewidth]{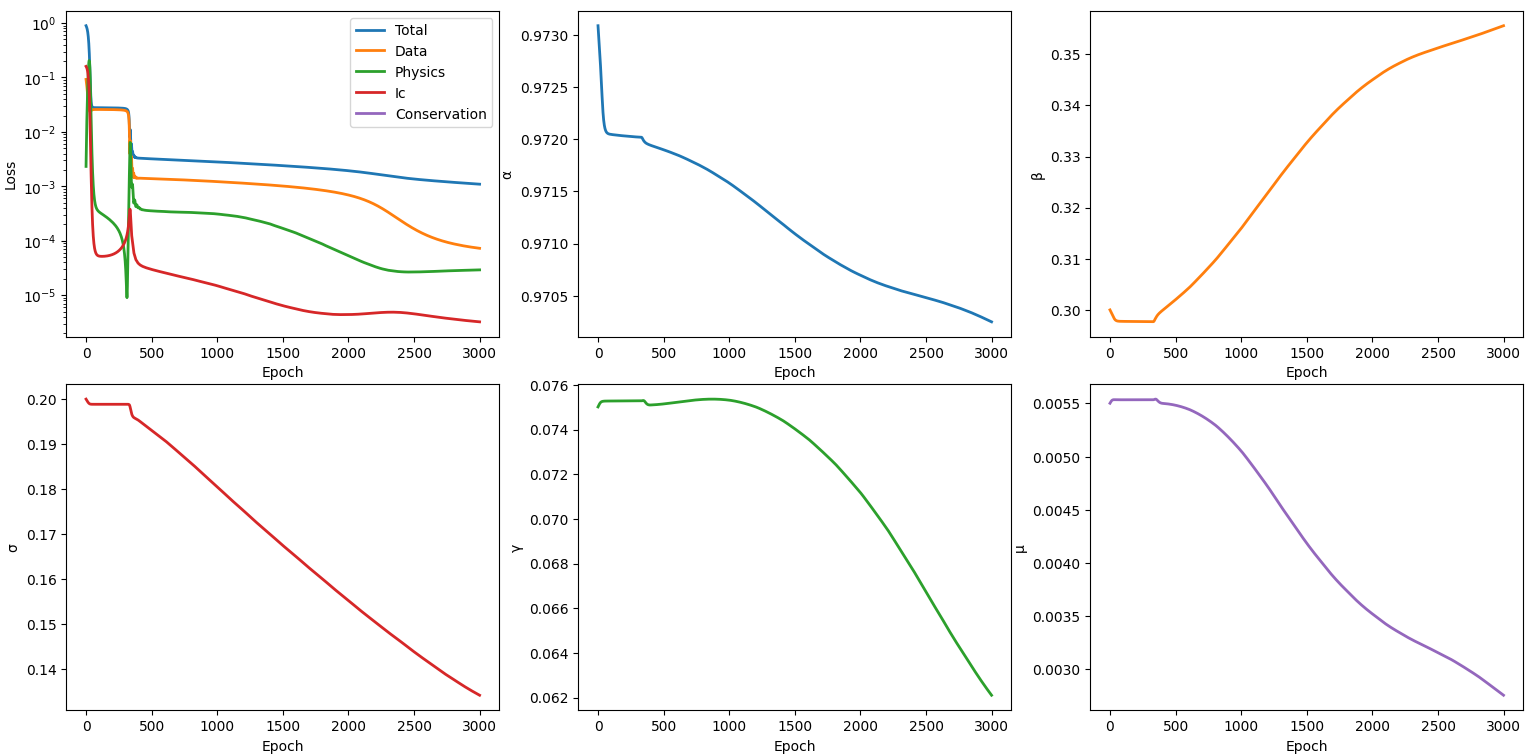}
\caption{Sweden data}
\end{subfigure}
\caption{Estimated epidemiological parameters for real data.}\label{F5}
\end{figure}

Figures~\ref{F4} and~\ref{F5} present the application of the proposed fractional-SEIRD PINN framework to real-world COVID-19 data from Germany and Sweden. 
Figure~\ref{F4} compares the reconstructed epidemic dynamics with the reported data for both countries. 
The trajectories of the susceptible, exposed, infected, recovered, or deceased populations show close agreement with the observations. This indicates that the PINN can capture the underlying transmission dynamics despite data noise and reporting variability. 
Notably, the German outbreak exhibits a sharper infection peak and a faster decline, while the Swedish epidemic has a more prolonged infectious curve. These differences are consistent with the different public health interventions implemented in the two countries.

Figure~\ref{F5} illustrates the corresponding parameter estimation outcomes. 
The recovered values remain within the admissible ranges provided in Table~\ref{Tab3}, confirming both the validity of the training process and the biological plausibility of the estimates. 
The fractional order $\alpha$ was consistently identified as less than one for both Germany and Sweden, reflecting the presence of long-memory effects in the epidemic data. 
Additionally, subtle differences emerge across the two case studies: the Swedish data suggest a slightly lower $\alpha$, accompanied by higher incubation and recovery rates, whereas the German data yield higher transmission and mortality rates. 
These contrasts demonstrate how fractional dynamics can reveal country-specific heterogeneity in epidemic progression that standard integer-order models may miss. 

Table~\ref{Tab4} summarizes the estimated parameters for Germany and Sweden.
All recovered values fall within the biologically plausible ranges reported in Table~\ref{Tab3}, confirming the reliability of the framework when applied to real data. 

\begin{table}[H]
\centering
\setlength{\tabcolsep}{0.5cm}
\caption{Estimated epidemiological parameters for COVID-19.}\label{Tab4}
\adjustbox{max width=\textwidth}{
\begin{tabular}{c||ccccc}
\hline
\textbf{Parameter} & $\alpha$ & $\beta$ & $\sigma$ & $\gamma$ & $\mu$ \\
\hline\hline
\textbf{Germany} & 0.915016 & 0.305806 & 0.274704 & 0.059172 & 0.002286 \\
\hline
\textbf{Sweden} & 0.970250 & 0.355629 & 0.134239 & 0.062103 & 0.002757 \\
\hline
\end{tabular}
}
\end{table}

Overall, the results confirm that the proposed PINN framework generalizes well to real-world COVID-19 data, providing both accurate trajectory reconstruction and interpretable parameter estimates. 
The identification of $\alpha < 1$ reinforces the epidemiological relevance of fractional-order models in accounting for memory effects such as delayed reporting, behavioral adaptation, and variability in disease progression.


\subsection{Limitations of the PINN Training}\label{S5.3}

Although the proposed fractional-SEIRD PINN framework yields promising results on synthetic and real datasets, it has several limitations. 
 
First, the framework relies on the availability and quality of epidemiological data. 
For real-world COVID-19 datasets, reporting delays, under-reporting, and missing recovery information can introduce uncertainties that affect the accuracy of reconstructed trajectories. 

Estimating the susceptible and exposed compartments is inherently challenging because they are usually unobserved and must be inferred indirectly. 
While the PINN provides a principled way to address this issue, further refinements could enhance robustness.

These limitations do not undermine the validity of the present results. Rather, they highlight areas for methodological improvement.

\section{Conclusion}\label{S6}

In this work, we developed a PINN framework for the fractional SEIRD epidemic model. Our focus was on estimating the fractional memory order $\alpha$ alongside classical epidemiological parameters. 
Leveraging the Caputo fractional derivative and the L1 discretization scheme, our method incorporates memory effects into epidemic dynamics, and captures long-range dependencies that are absent in classical integer-order formulations. 
Validation on Mpox synthetic datasets demonstrated that the proposed framework can accurately recover $\alpha$ and epidemiological parameters, even in the presence of observational noise. 
Applications to real datasets further confirmed that $\alpha < 1$, supporting the hypothesis that epidemic spreading processes exhibit non-Markovian behavior. 
Let us note that a fractional derivative with order $\alpha < 1$ implies that system dynamics depend on the entire history of the process, not just the present state. 
This nonlocality captures the heavy tails of the incubation and infectious periods, reflecting empirical evidence that epidemic processes exhibit memory effects that are inconsistent with purely Markovian (exponential waiting time) assumptions. 
A comparative analysis against the classical SEIRD model revealed clear improvements in predictive accuracy and interpretability when memory was incorporated.  
Overall, our findings suggest that the fractional SEIRD model coupled with PINNs is a powerful tool for epidemic analysis, offering improved predictions and deeper insights into the underlying dynamics of infectious diseases.

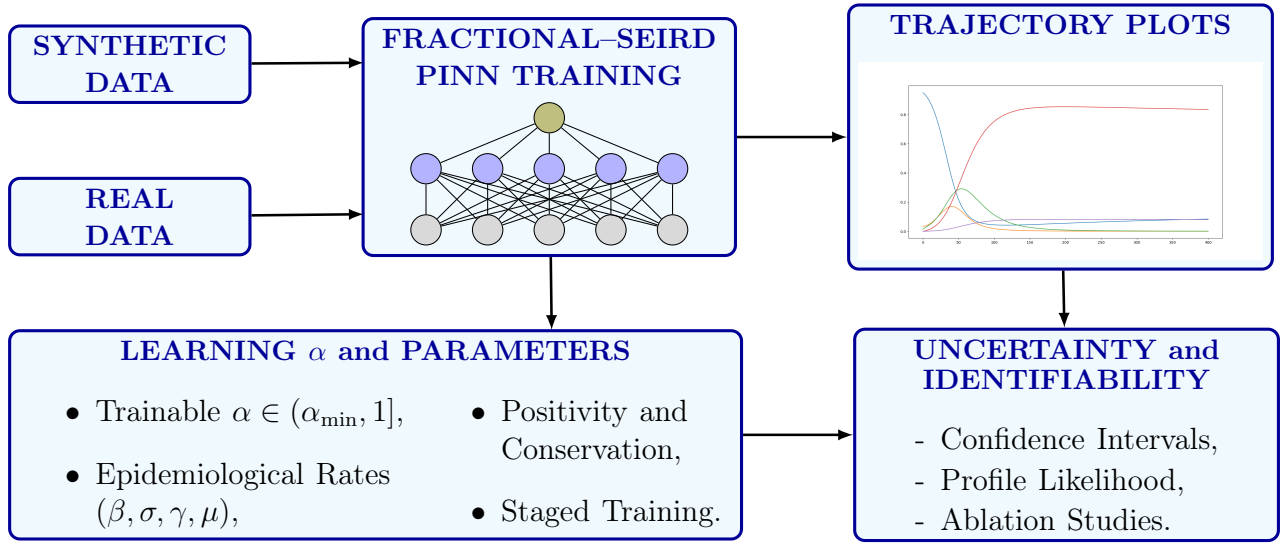
\begin{figure}[H]
\centering
\adjustbox{max width=\textwidth}{
\begin{tikzpicture}[node distance=3cm]
\node (SD) [draw=blue!60!black, very thick, rounded corners,
font=\bfseries, text width=3cm, fill=cyan!5, align=center] 
{\small\centering{\color{blue!60!black}SYNTHETIC DATA}};
\node (RD) [draw=blue!60!black, very thick, rounded corners,
font=\bfseries, text width=3cm, fill=cyan!5, align=center, below = of SD, yshift=2cm] 
{\small\centering{\color{blue!60!black}REAL\\ DATA}};
\node (Training) [draw=blue!60!black, very thick, rounded corners,
font=\bfseries, text width=4.8cm, fill=cyan!5, align=center, right = of SD, yshift=-1cm, xshift=-1.5cm] 
{\small\centering{\color{blue!60!black}FRACTIONAL--SEIRD PINN TRAINING\vspace{0.2cm}}
\begin{tikzpicture}[scale=0.7]
\node[circle, draw, fill=green!50!red!50, minimum size=0.4cm] (i0) at (2.4, 0) {};
\node[circle, draw, fill=blue!30, minimum size=0.4cm] (i1) at (0, -1) {};
\node[circle, draw, fill=blue!30, minimum size=0.4cm] (i2) at (1.2, -1) {};
\node[circle, draw, fill=blue!30, minimum size=0.4cm] (i3) at (2.4, -1) {};
\node[circle, draw, fill=blue!30, minimum size=0.4cm] (i4) at (3.6, -1) {};
\node[circle, draw, fill=blue!30, minimum size=0.4cm] (i5) at (4.8, -1) {};
\node[circle, draw, fill=gray!30, minimum size=0.4cm] (i6) at (0, -2.2) {};
\node[circle, draw, fill=gray!30, minimum size=0.4cm] (i7) at (1.2, -2.2) {};
\node[circle, draw, fill=gray!30, minimum size=0.4cm] (i8) at (2.4, -2.2) {};
\node[circle, draw, fill=gray!30, minimum size=0.4cm] (i9) at (3.6, -2.2) {};
\node[circle, draw, fill=gray!30, minimum size=0.4cm] (i10) at (4.8, -2.2) {};
\foreach \i in {i1,i2,i3,i4,i5}
\foreach \j in {i0}
\draw (\i) -- (\j);
\foreach \i in {i1,i2,i3,i4,i5}
\foreach \h in {i6,i7,i8,i9,i10}
\draw (\i) -- (\h);
\end{tikzpicture}
};
\node (Plots) [draw=blue!60!black, very thick, rounded corners,
font=\bfseries, text width = 5.5cm, fill=cyan!5, right = of Training, xshift=-1.5cm] 
{\small\centering{\color{blue!60!black}TRAJECTORY PLOTS} \footnotesize
\begin{equation*}
\includegraphics[width=1\textwidth]{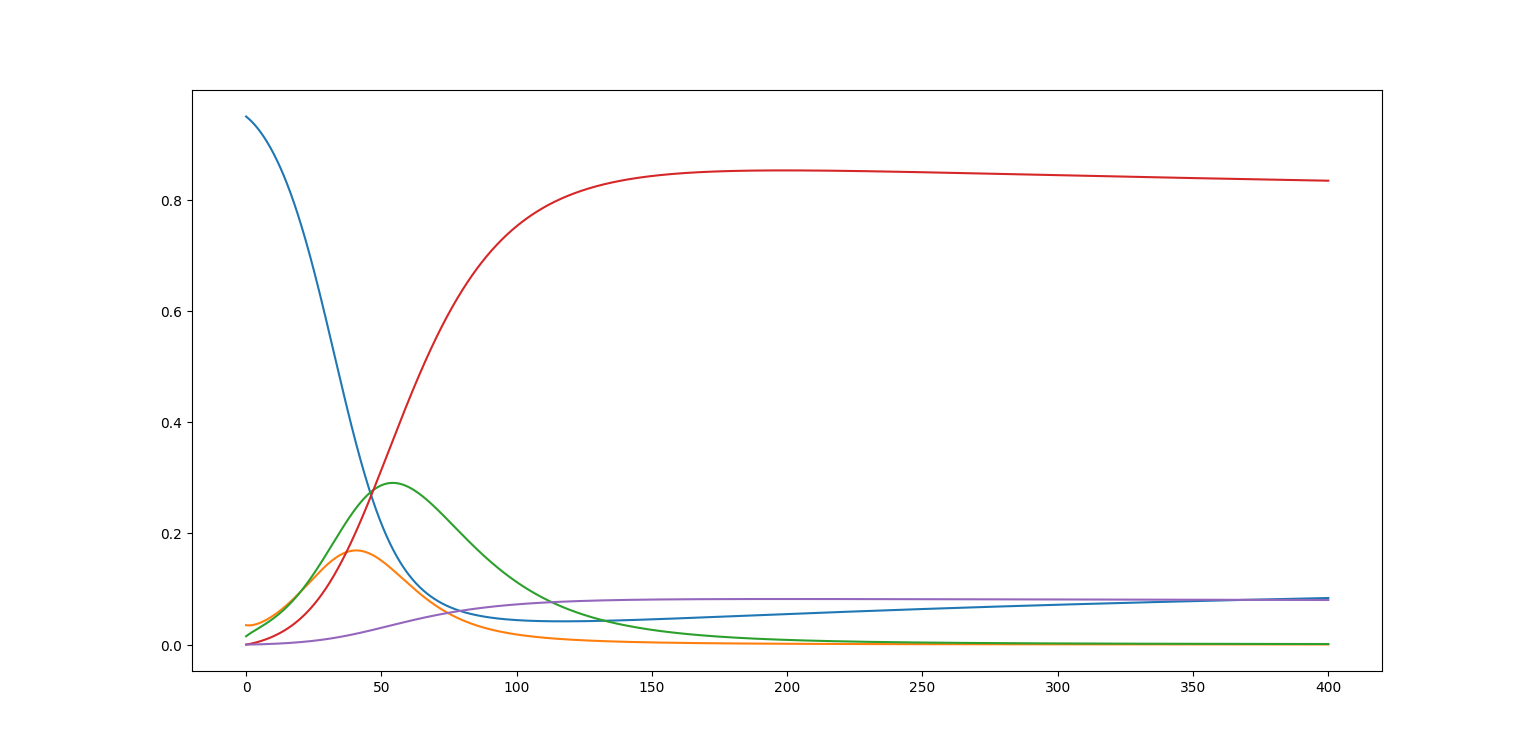}
\end{equation*}
};
\node (Learning) [draw=blue!60!black, very thick, rounded corners,
font=\bfseries, text width=9.7cm, fill=cyan!5, align=center, below = of RD, yshift=2cm, xshift=3.35cm] 
{\small\centering{\color{blue!60!black}LEARNING $\alpha$ and PARAMETERS}\footnotesize

\raggedright{
\begin{table}[H]
\begin{minipage}{0.55\linewidth}
\begin{itemize}
\item Trainable $\alpha \in (\alpha_{\min},1]$,

\item Epidemiological Rates ($\beta, \sigma, \gamma, \mu$),
\end{itemize}
\end{minipage}
\hfill
\begin{minipage}{0.43\linewidth}
\begin{itemize}
\item Positivity and Conservation,

\item Staged Training.
\end{itemize}
\end{minipage}
\end{table}
}};
\node (Unc) [draw=blue!60!black, very thick, rounded corners,
font=\bfseries, text width = 5.5cm, fill=cyan!5, right = of Learning, xshift=-1.5cm] 
{\small\centering{\color{blue!60!black}UNCERTAINTY and IDENTIFIABILITY} \footnotesize

\raggedright{
\begin{table}[H]
\begin{minipage}{\linewidth}
\begin{itemize}
\item[-] Confidence Intervals,
\vspace{-0.3cm}
\item[-] Profile Likelihood,
\vspace{-0.3cm}
\item[-] Ablation Studies.
\end{itemize}
\end{minipage}
\end{table}
}};
\draw [->, thick, >=latex, line width=1pt] (SD) -- (3.19,0.02);
\draw [->, thick, >=latex, line width=1pt] (RD) -- (3.19,-2.06);
\draw [->, thick, >=latex, line width=1pt] (Training) -- (5.75,-3.65);
\draw [->, thick, >=latex, line width=1pt] (Training) -- (Plots);
\draw [->, thick, >=latex, line width=1pt] (Plots) -- (Unc);
\draw [->, thick, >=latex, line width=1pt] (Learning) -- (Unc);
\end{tikzpicture}
}
\captionof{figure}{Workflow of PINNs in the Fractional SEIRD Model.}\label{F6}
\end{figure}

Figure~\ref{F6} illustrates the overall workflow of the proposed approach. Both synthetic and real datasets serve as inputs to the fractional SEIRD model, which uses a PINN architecture to maintain dataset fidelity and ensure physical consistency. 
The model outputs compartmental trajectories, from which key quantities such as the fractional order $\alpha$ and epidemiological parameters are inferred. The schematic also illustrates the importance  of conservation constraints and uncertainty analysis inensuring, that the learned parameters are interpretable and reliable. 
The diagram emphasizes the integration of data-driven learning and fractional epidemic modeling into a unified framework. 

This study reveals several promising directions. 
One natural extension is introducing a time-varying fractional order $\alpha(t)$, which would enable the model to capture evolving memory effects during different epidemic phases, such as periods of intervention or changes in population behavior. 
Another important avenue is integrating multi-region and network-based epidemic models. 
Spatial coupling and mobility data could be incorporated to infer $\alpha$ across interconnected regions, improving large-scale epidemic forecasting. 
Developing spatiotemporal extensions of the fractional SEIRD model within the PINN framework would be particularly appealing in this context because it would enable the simultaneous representation of memory effects, temporal evolution, and spatial heterogeneity in epidemic dynamics.

In terms of uncertainty quantification, shifting from bootstrap-based methods to Bayesian formulations, such as Bayesian neural operators or variational inference, could yield more reliable posterior distributions for $\alpha$ and epidemiological parameters. 
Improving scalability and efficiency is essential on the computational side, especially by designing fast algorithms for fractional derivative evaluation, such as reduced-memory kernels or GPU-accelerated schemes, to make the framework practical for real-time epidemic monitoring. 
Finally, applying the approach to other infectious diseases, such as seasonal influenza, dengue, and vector-borne epidemics, in addition to Mpox and COVID-19, would demonstrate its generality and highlight the role of long-memory effects in different epidemiological contexts. 

In summary, this work establishes a foundation for integrating fractional dynamics with PINNs in epidemic modeling. By demonstrating the identifiability and usefulness of the fractional memory order $\alpha$, it paves the way for richer, spatiotemporal, memory-aware epidemic models that can improve scientific understanding and public health decision-making.

\section*{Acknowledgments}

The author would like to express sincere thanks to \emph{Dr.\ Rachid Benoudi}, \emph{Prof.\ Dr.\ Matthias Ehrhardt}, and \emph{Prof.\ Dr.\ Moulay Rchid Sidi Ammi} for 
their valuable feedback and guidance.



\section*{Declarations}

\subsection*{Data Availability} 

This study uses both synthetic and real epidemic datasets. Real datasets were obtained from the following publicly available source: \cite{OWD2025, Gehrcke2025, OWD2025Sweden, worldometers}.

\subsection*{Conflict of Interest} 

The author declares that no conflict of interest exists that could have influenced the study presented in this paper.


\bibliographystyle{unsrt}
\bibliography{References}

\end{document}